\pdfoutput=1
\documentclass[11pt]{article}
\usepackage{emnlp2021}
\usepackage{times}
\usepackage{svg}
\usepackage{latexsym}
\usepackage{tabularx}
\usepackage{pgfplots} 
\pgfplotsset{compat=1.17} 
\usepackage[T1]{fontenc}
\usepackage[utf8]{inputenc}
\usepackage{microtype}
\usepackage{multirow}
\usepackage{rotating}
\usepackage{pdflscape} % for 'landscape' environment
\usepackage{longtable}
\usetikzlibrary{patterns}
\usepackage[normalem]{ulem}
\useunder{\uline}{\ul}{}

\title{MFAQ: a Multilingual FAQ Dataset}

\author{
    Maxime De Bruyn, Ehsan Lotfi, Jeska Buhmann, Walter Daelemans \\
    CLiPS Research Center \\
    University of Antwerp, Belgium \\
    \texttt{firstname.lastname@uantwerpen.be}
}

\begin{document}
\maketitle
\begin{abstract}
In this paper, we present the first multilingual FAQ dataset publicly available. We collected around 6M FAQ pairs from the web, in 21 different languages. Although this is significantly larger than existing FAQ retrieval datasets, it comes with its own challenges: duplication of content and uneven distribution of topics. 
We adopt a similar setup as Dense Passage Retrieval (DPR) \cite{karpukhin-etal-2020-dense} and test various bi-encoders on this dataset. Our experiments reveal that a multilingual model based on XLM-RoBERTa \cite{DBLP:journals/corr/abs-1911-02116} achieves the best results, except for English. Lower resources languages seem to learn from one another as a multilingual model achieves a higher MRR than language-specific ones. Our qualitative analysis reveals the brittleness of the model on simple word changes. We publicly release our dataset\footnote{\href{https://huggingface.co/datasets/clips/mfaq}{https://huggingface.co/datasets/clips/mfaq}}, model\footnote{\href{https://huggingface.co/clips/mfaq}{https://huggingface.co/clips/mfaq}} and training script\footnote{\href{https://github.com/clips/mfaq}{https://github.com/clips/mfaq}}.
\end{abstract}

\section{Introduction}

Organizations create \emph{Frequently Asked Questions} (FAQ) pages on their website to provide a better service to their users. FAQs are also useful to automatically answer the most frequent questions on different communication channels: email, chatbot, or search bar.

FAQ retrieval is the task of locating the right answer within a  collection of candidate question and answer pairs. It is closely related to the tasks of \emph{non-factoid QA} and \emph{community QA}, although it has its own specificities. The total number of possible answers is generally small (the average FAQ page on the web has 6 answers), and only one is correct. Retrieval systems cannot rely on named entities, as they are typically shared by many possible answers. For example, three out of four answers in Table \ref{example-dataset} share the \emph{COVID-19} entity. Lastly, new user queries are matched against pairs of questions and answers, as opposed to passages for non-factoid QA.

\begin{table}
\begin{tabularx}{\columnwidth}{X}
\multicolumn{1}{c}{\textbf{Example FAQs}}                                                                                              \\ \hline
\textit{Is it safe for my child to get a COVID-19 vaccine?} Yes. Studies show that COVID-19 vaccines are safe and effective. {[}...{]}                   \\ \hline
\textit{If I am pregnant, can I get a COVID-19 vaccine?} Yes, if you are pregnant, you can receive if COVID-19 vaccine.                         \\ \hline
\textit{What are the ingredients in COVID-19 vaccines?} Vaccine ingredients can vary by manufacturer.                                 \\ \hline
\textit{How long does protection from a COVID-19 vaccine last?} We don't know how long protection lasts for those who are vaccinated. {[}...{]} \\ \hline
\end{tabularx}
\caption{Example FAQs about the COVID-19 vaccine from the CDC website.}
\label{example-dataset}
\end{table}

Since FAQ-Finder \cite{hammond_faq_1995}, researchers applied different methods to the task of FAQ retrieval \cite{sneiders_automated_1999, jijkoun_retrieving_2005, riezler_statistical_2007, 10.1007/978-3-319-45510-5_9, DBLP:journals/corr/abs-1905-02851}. However, since the advent of deep-learning and Transformers, the interest has somewhat faded compared to other areas of QA \cite{DBLP:journals/corr/abs-2107-12708}. One possible explanation is the lack of a dedicated large-scale dataset. The ones available are mostly limited to English, and domain-specific.

On the other hand, the task of factoid question answering received the attention of many researchers. Recently, Transformers encoders such as Dense Passage Retrieval (DPR) \cite{karpukhin-etal-2020-dense} have been successfully applied to the retrieval part of factoid QA, overcoming strong baselines such as TF-IDF and BM25. However, we show that DPR's performance on passage retrieval is not directly transferable to FAQ retrieval. \citet{DBLP:journals/corr/abs-2102-07033} recently released PAQ, a dataset of 65M pairs of \emph{Probably Asked Questions}. However, answers are typically short in PAQ (a few words), which differs from FAQs where answers are longer than questions.

Another way to answer users' questions is to use \emph{Knowledge Grounded Conversation} models as it does not require the pre-generation of all possible pairs of questions and answers \cite{DBLP:journals/corr/abs-2107-07566, de2020bart}. However, at the time of writing these models can hallucinate knowledge \cite{DBLP:journals/corr/abs-2104-07567}, which limits their attractiveness in a corporate environment.

In this paper, we provide the first multilingual dataset of FAQs. We collected around 6M FAQ pairs from the web in 21 different languages. This is significantly larger than existing datasets. However, collecting data from the web brings its own challenges: duplication of content and uneven distribution of topics. We also provide the first multilingual FAQ retriever. We show that models trained on all languages at once outperform monolingual models (except for English).

The remainder of the paper is organized as follows. We first review the existing models and datasets available for the task of FAQ retrieval. We then present our own dataset and apply different models to it. We finally perform some analysis on the results and conclude. Our dataset and model are available on the HuggingFace Hub\footnote{\href{https://huggingface.co/datasets/clips/mfaq}{dataset}, \href{https://huggingface.co/clips/mfaq}{model} and \href{https://github.com/clips/mfaq}{training script}}.

\section{Related Work}
In this section, we review the existing literature on FAQ retrieval. We first start by reviewing available models and then look at the available datasets.

\subsection{Models}
Since the release of FAQ-Finder \cite{hammond_faq_1995, burke_question_1997} and Auto-FAQ \cite{whitehead_auto-faq_1995}, several methods have been presented. We grouped them into three categories: lexical, unsupervised, and supervised.

\paragraph{Lexical}
FAQ-Finder \cite{hammond_faq_1995, burke_question_1997} matches user queries to FAQ questions of the Usenet dataset using Term Frequency-Inverse Document Frequency (TF-IDF). The system tries to bridge the lexical gap between users' queries and FAQ pairs by using the semantic network \emph{WordNet} \cite{miller1995wordnet} to establish correlations between related terms. FAQ-Finder assumes that the question half of the QA pair is the most relevant for matching to a new query. \citet{tomuro_retrieval_2004} improved upon FAQ-Finder by including the other half of the QA pair (the answer).
\citet{xie_faq-based_2020} uses a knowledge graph-based QA framework that considers entities and triples in texts as knowledge anchors. This approach requires the customization of a knowledge graph, which is labor-intensive and domain-specific.

\citet{sneiders_automated_1999} used a rule-based technique called Prioritized Keyword Matching on top of a traditional TF-IDF approach. The use of shallow language understanding means that the matching is based on keyword comparison. Each FAQ entry must be manually annotated with a set of required and optional keywords. \citet{sneiders_automated_2002-1, sneiders_automated_2009, sneiders_automated_2010} brings further developments on the idea. \citet{moreo_learning_2013} proposes an approach based on semi-automatic generation of regular expression for matching queries with answers. \citet{yang_developing_2009} integrates a domain ontology, user modeling, and a template-based approach to tackle this problem.

\paragraph{Unsupervised}
\citet{kim_cluster-based_2008, kim_high-performance_2006} presented a clustering-based method of previous user queries to retrieve the right FAQ pair. The authors used a Latent Semantic Analysis (LSA) method to overcome the lexical mismatch between related queries.
\citet{jijkoun_retrieving_2005} experimented with several combinations of TF-IDF retrievers based on the indexing of different fields (question, answer, with or without stop words, the full text of the page). \citet{riezler_statistical_2007}
  extended this method by incorporating a translation-based query expansion, as initially investigated in \citet{berger_bridging_2000}.

\paragraph{Supervised}
\citet{moschitti_exploiting_2007} proposed an approach based on tree kernels. Tree kernels can be defined as similarity metrics that compare a query to an FAQ pair by parsing both texts and calculating the similarity based on the resulting parse trees. Semantic word similarity can also be added to the computation. \citet{filice_kelp_2016} expanded on this method and achieved first place in the Community QA shared task at SemEval 2015 \cite{nakov-etal-2015-semeval}.

\citet{DBLP:journals/corr/abs-1905-02851} were the first to use BERT-based models \cite{DBLP:journals/corr/abs-1810-04805} for the specific task of FAQ retrieval. The relevance between the query and the answers is learned with a fine-tuned BERT model which outputs probability scores for a pair of (query, answer). The scores are then combined using a specific method. \citet{mass-etal-2020-unsupervised} also used a BERT model. Their method is based on an initial retrieval of FAQ candidates followed by three re-rankers. \citet{DBLP:journals/corr/abs-2108-00719} used a ConveRT \cite{DBLP:journals/corr/abs-1911-03688} model to automatically answer FAQ questions in Dutch.

\subsection{Datasets}

\begin{table*}
\centering
    \resizebox{\textwidth}{!}
    {
        \begin{tabular}{lllllll}
        \hline
        \textbf{Name} & \textbf{Size} & \textbf{Lang.} & \textbf{Domain}        & \textbf{Source}    & \textbf{Q\textgreater{}1} & \textbf{A\textgreater{}1} \\ \hline
        Usenet \citep{hammond_faq_1995}        & -               & En             & Multi-domain           & Usenet             & No                        & No                        \\
        FAQIR \citep{10.1007/978-3-319-45510-5_9}          & 4,313         & En             & Maintenance & Yahoo! Answers     & Yes                       & Yes                       \\
        StackFAQ \citep{karan_paraphrase-focused_2018}       & 719           & En             & Web apps               & StackExchange      & Yes                       & Yes                       \\
        InsuranceQA \citep{feng_applying_2015}    & 12,887        & En             & Insurance              & Insurance Library  & No                        & Yes                       \\
        CQA-QL \citep{nakov-etal-2015-semeval}       & 2,600         & En             & Qatar                  & Qatar living forum & No                        & Yes                       \\
        Fatwa corpus \citep{nakov-etal-2015-semeval} & 1,300         & Ar             & Quran                  & Fatwa website      & No                        & Yes                       \\ \hline
        M-FAQ (ours) & 6,134,533         & Multi             & Multi                 & Multi      & No                        & No                       \\ \hline
        \end{tabular}
    }
    \caption{
    List of the common datasets used in FAQ retrieval. Size is the number of pairs available. Q>1 denotes if the dataset has multiple available questions for a single answer (i.e., does the dataset have paraphrases), while A>1 denotes if the dataset has multiple answers for a given question.
    }
    \label{all-datasets}
\end{table*}

In this section, we review the different datasets publicly available. FAQ retrieval datasets can be evaluated on four axes: source of data (community or organizational), the existence of user queries (paraphrases), domain, and language. See Table \ref{all-datasets} for an overview.

Faq-Finder \cite{hammond_faq_1995, burke_question_1997} used a dataset collected from Usenet news groups. FAQs were created on several topics so that newcomers do not ask the same questions again and again. This dataset is multi-domain. More recently, \citet{10.1007/978-3-319-45510-5_9} released the FAQIR dataset. It was collected from the "maintenance \& repairs" section of the QA website \emph{Yahoo! Answers}. The StackFAQ \cite{karan_paraphrase-focused_2018} dataset was collected from the "web apps" sections of \emph{StackExchange}. \citet{feng_applying_2015} collected a QA dataset from the  \href{http://www.insurancelibrary.com}{insurancelibrary.com} website where a community of insurance expert reply to users' questions. Several authors (for example \citealp{filice_kelp_2016}) also rely on Sem-Eval 2015 Task 3 \cite{nakov-etal-2015-semeval} on Answer Selection in Community Question Answering. It contains pairs of questions and answers in English and Arabic.

There exist few publicly available datasets for organizational FAQs. OrgFAQ \cite{DBLP:journals/corr/abs-2009-01460} is a notable exception. At the time of writing, it is not yet publicly available. 

\section{Multilingual FAQ dataset}
In this section, we introduce our new multilingual FAQ dataset.

\subsection{Data collection}
Instead of implementing our own web crawler, we used the Common Crawl: a non-profit organization which provides an open repository of the web.\footnote{https://commoncrawl.org/about/} Common Crawl's complete web archive consists of petabytes of data collected over 10 years of web crawling \cite{ortiz-suarez-etal-2020-monolingual}. The repository is organized in monthly bucket of crawled data. Web pages are saved in three different formats: WARC files for the raw HTML data, WAT files for the metadata, and WET files for the plain text extracts.

For our purposes, we used WARC files as we are interested in the raw HTML data. Similar to \citet{DBLP:journals/corr/abs-2009-01460}, we looked for \emph{JSON-LD}\footnote{JavaScript Object Notation for Linked Data} tags containing an \emph{FAQPage} item. Web developers use this tag to make it easy for search engines to parse FAQs from a web page.\footnote{More information on \href{https://developers.google.com/search/docs/advanced/structured-data/faqpage}{FAQPage} markup} The language of each FAQ pair is determined with fastText \cite{DBLP:journals/corr/JoulinGBM16}. We also apply some filtering to remove unwanted noise.\footnote{Questions need to contain a question mark (including the Arabic question mark) to avoid keyword questions. Question and answer cannot start with a "<", "\{", or "[" to remove "code like" data.} Using this method, we collected 155M FAQ pairs from 24M different pages.

\subsection{Deduplication}
A common issue with datasets collected from the web is the redundancy of data \cite{DBLP:journals/corr/abs-2107-06499}.
For example, hotel pages on \emph{TripAdvisor} typically have an FAQ pair referring to shuttle services from the airport to the hotel.\footnote{Does Ritz Paris have an airport shuttle? Does Four Seasons Hotel George V have an airport shuttle?} The only changing term is the name of the hotel.

Algorithms such as SimHash \cite{10.1145/509907.509965} and MinHash \cite{666900} can detect such duplicates. MinHash is an approximate matching algorithm widely used in large-scale
deduplication tasks \cite{DBLP:journals/corr/abs-2107-06499, versley2012not, gabriel2018identifying, DBLP:conf/lrec/GyawaliAK20}. The main idea of MinHash is to efficiently estimate the Jaccard similarity between two documents, represented by their set of n-grams. Because of the sparse nature of n-grams, computing the full Jaccard similarity between all documents is prohibitive. MinHash alleviates this issue by reducing each document to a fixed-length hash which can be used to efficiently approximate the Jaccard similarity between two documents. MinHash has the additional property that similar documents will have similar hashes, we can then use Locality Sensitive Hashing (LSH) \cite{10.5555/2787930} to efficiently retrieve similar documents.

In our experiments, we represented each page as a set of 3 consecutive tokens (n-grams). We worked with a document signature length of 100, and 20 bands with 5 rows as parameters for LSH. These parameters ensure a 99.6\% probability that documents with a Jaccard similarity of 0.75 will be identified. We subsequently compute the true Jaccard similarity for all matches.

We follow the approach of \emph{NearDup} \cite{DBLP:journals/corr/abs-2107-06499} and subsequently create a graph of documents. Each node on the graph is an FAQ page, and they share an edge if their true Jaccard similarity is larger than 0.75. We then compute all the independent sub-graphs, each representing a graph of duplicated pages. We only keep one page per sub-graph.

Using this method, we trimmed the number of FAQ pages from 24M to 1M.

\subsection{Description}
After deduplication, our dataset contains around 6M FAQ pairs coming from 1M different web pages, spread on 26K root web domains.\footnote{We define a root web domain as the last substring before the extension (e.g. TripAdvisor is the root web domain in fr.tripadvisor.com). In other words, we strip the extension and any subdomain.} This is significantly bigger than other FAQ datasets publicly available at the time of writing (see Table \ref{all-datasets} for comparison).

Our dataset is composed of pairs of FAQs grouped by language and source page (URL). We collected data in 21 different languages.\footnote{We did not target specific languages, however, we removed languages with fewer than 250 pairs. Common languages such as Chinese, Hindi, Arabic and Japanese are missing. Although we do not have an official reason why, we think it may be because of our initial filtering or the fact ldjson markup is not widely used in these languages.} The most common one is English, with 58\% of the FAQ pairs, followed by German and Spanish with 13\% and 8\% respectively. 

\subsection{Training and validation sets}
For a given language, the target size of the validation set is equal to 10\% of the total number of pairs. However, two features of our dataset call for a more fine-grained approach. 

\subsubsection{Root domain distribution}
Even though we deduplicated the dataset, FAQ pages tend to originate from the same root domain. As an example, kayak (\emph{kayak.com}, \emph{kayak.es}, etc.) is the largest contributor to the dataset. While this is not a problem for the training set (one can always restrict the number of pages per domain), it is an issue for the validation set, as we want to assess the quality of the model on a broad set of topics. Having several large root domain contributors skews the dataset to these topics. We make the simplifying assumption that different web domains have different topics of interest. Research on the true topic distribution is left for future work.

We artificially increased the topic breadth of the validation set by restricting the contribution of each root domain. In the validation set, a single root domain can only contribute up to 3 FAQ pages. This method reduces the contribution of the largest domain from 21\% in the training set to 3\% in the validation set. Furthermore, we make sure there is no overlap of root domain between the training and validation set.\footnote{We use the root domain instead of the regular domain name to avoid having \emph{help.domain.com} in the training set and \emph{domain.co.uk} in the validation set}

% See Figure \ref{root-domain-coverage} for a comparison between the training and validation set. 

\subsubsection{Pairs per page concentration}
\begin{figure}
\centering
\resizebox{\columnwidth}{!}
{
    \begin{tikzpicture}
        \begin{axis}[
            % xtick style={draw=none},
            symbolic x coords={5,10,15,20,25,30,30+},
            %ylabel=Percentage in bucket,
        	% enlargelimits=0.15,
        	yticklabel={$\pgfmathprintnumber{\tick}\%$},
            xlabel={Bins},
            ylabel={Coverage},
        	legend style={anchor=north east},
        	ytick={0,20,40,60,80,100},
        	ymin=0, ymax=110,
        	ybar=4pt,% configures `bar shift'
        	%ymajorgrids=true,
            %grid style={line width=0.1pt,},
            grid = both,
            grid style={dashed, line width=0.1pt, gray!33},
        	bar width=6pt,
        	xtick align=inside,
        	]
            \addplot[ybar,fill=blue!30!white,postaction={pattern=north west lines}] coordinates {
                (5,40)
                (10,48)
                (15,9)
                (20,2)
                (25,0)
                (30,0)
                (30+,0)
            };
            \addplot[ybar,fill=red!30!white,postaction={pattern=north east lines}] coordinates {
                (5,0)
                (10,39)
                (15,24)
                (20,13)
                (25,8)
                (30,4)
                (30+,11)
            };
            \legend {Training, Validation};
        \end{axis}
    \end{tikzpicture}
}
    \caption{
        Bucketing of our dataset according to the number of FAQs per page. To make the validation set more challenging, we started by selecting pages with a higher number of pairs.
    }
    \label{pairs-per-page}
\end{figure}
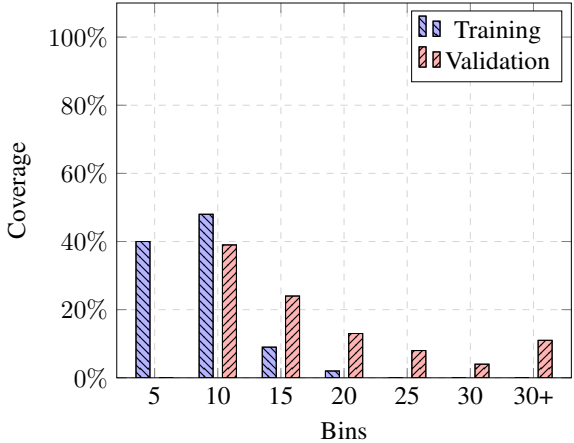

The distribution of the number of pairs per page is highly uneven (see Figure \ref{pairs-per-page}). Around 50\% of the pages have 5 or fewer pairs per page. Intuitively, we prefer pages with a higher number of FAQs as it is harder to pick the right answers amongst 100 candidates than 5. We thus artificially increased the difficulty of the validation by first selecting pages with a higher number of FAQ pairs per page. See Figure \ref{pairs-per-page} for a comparison between the training and validations set.

\subsubsection{Cross-lingual leakage}
The fact that our dataset is multilingual can lead to issues of cross-lingual leakage. Having pages from \emph{expedia.fr} in the training set, and pages from \emph{expedia.es} in the validation set can overstate the performance of the models. We avoid such problems by restricting root domains in the validation associated with only one language (e.g. \emph{expedia} would be excluded from the validation set because it is associated with French and Spanish pages).

\begin{table}
\resizebox{\columnwidth}{!}
{
\begin{tabular}{lrrr}
\hline
\textbf{Language} & \textbf{Pairs}   & \textbf{Pages}     & \textbf{Domains} \\ \hline
English       & 3,719,484 & 608,796  & 17,635   \\
German        & 829,098 & 117,618   & 2,948   \\
Spanish       & 482,818 & 75,489   & 1,610   \\
French        & 351,458 & 56,317   & 1,795   \\
Italian       & 155,296 & 24,562  & 685     \\
Dutch         & 150,819 & 32,574   & 1,472     \\
Portuguese    & 138,778 & 26,169    & 608     \\
Turkish       & 102,373 & 19,002   & 580     \\
Russian       & 91,771 & 22,643    & 953     \\
Polish        & 65,182 & 10,695     & 445     \\
Indonesian    & 45,839  & 7,910     & 309     \\
Norwegian     & 37,711  & 5,143    & 198     \\
Swedish       & 37,003 & 5,270      & 434     \\
Danish        & 32,655  & 5,279     & 362     \\
Vietnamese    & 27,157 & 5,261      & 469     \\
Finnish       & 20,485 & 2,795      & 234     \\
Romanian      & 17,066  & 3,554    & 152     \\
Czech         & 16,675  & 2,568     & 182     \\
Hebrew        & 11,212  & 1,921   & 205     \\
Hungarian     & 8,598   & 1,264     & 150     \\
Croatian      & 5,215   & 819     & 99      \\ \hline
Total         & 6,346,693 & 1,035,649 & 31,525 \\ \hline
\end{tabular}
}
    \caption{
    Summary statistics about our dataset.
    }
    \label{table-short-datasets}
\end{table}

\section{Models}
In this section, we describe the FAQ retrieval models used in our experiments. Let \(P\) be the set of all user queries and \(F = \{(q_1, a_1), ..., (q_n, a_n)\}\) be the set of all FAQ pairs for a given domain. An FAQ retrieval model takes as input a user’s query \begin{math}p_i \in P\end{math} and an FAQ pair \begin{math}f_j \in F\end{math}, and outputs a relevance score \(h(p_i, f_j)\) for \(f_j\) with respect to \(p_i\). However, our dataset does not contain live user queries (or paraphrases) \(P\), we thus use questions \(q\) as queries \(P = \{q_1, ... , q_n\}\) and restrict the FAQ set to the answers \(F = \{a_1, ..., a_n\}\). The task becomes to rank the answers \(A\) according to the questions \(Q\).

\subsection{Baselines}
We experimented with several baselines: two unsupervised and one supervised.
\subsubsection{TF-IDF}
The traditional information retrieval method \cite{salton1975vector} uses a vector representation  for \(q_i\) and \(a_i\) and computes a dot-product as similarity relevance score \(h(q_i, a_i)\). We use n-grams of size (1, 3) and fit one model per FAQ page.

\subsubsection{Universal Sentence Encoder}
Encoding the semantics of a question \(q_i\) and an answer \(a_i\) can be achieved with the Universal Sentence Encoder \cite{DBLP:journals/corr/abs-1803-11175}. The model works on monolingual and multilingual data. We encode each question and answer independently, and then perform a dot-product of the questions' and answers' representations.

\subsubsection{Dense Passage Retrieval (DPR)}
Dense Passage Retrieval (DPR) \cite{karpukhin-etal-2020-dense} is a state of the art method for passage retrieval. It uses a bi-encoder to encode questions and passages into a shared embedding space. We fine-tune DPR on our dataset using the same procedure described in Section \ref{section:training}.

\subsection{XLM-Roberta as bi-encoders}\label{xlm-roberta}
Bi-encoders encode questions \(q_i\) and answers \(a_i\) independently and output a fixed \emph{d}-dimensional representation for each query and answer. The encoder can be shared or independent to generate the representations.\footnote{We use a shared encoder, which means we use the same network to compute the representation for questions and answers. DPR uses independent encoders.} At run-time, new queries are encoded using the encoder, and the top-\emph{k} closest answers are returned. The representations for the answers can be computed once, and cached for later use. Similarity is typically computed using a dot product.

\subsubsection{Multilingual}
The state-of-the-art encoders such as RoBERTa \cite{DBLP:journals/corr/abs-1907-11692} and BERT \cite{DBLP:journals/corr/abs-1810-04805} are trained for English only. As our dataset is multilingual we opted for XLM-RoBERTa \cite{DBLP:journals/corr/abs-1911-02116}, it was trained using masked language modeling on one hundred languages, using more than 2TB of filtered CommonCrawl data. This choice allows us to leverage the size of the English data for less represented languages.

\subsubsection{Training} \label{section:training}
Given pairs of questions and answers, along with a list of non-relevant answers, the bi-encoder model is trained to minimize the negative log-likelihood of picking the positive answer amongst the non-relevant answers. Non-relevant answers can be divided into \emph{in-batch} negatives and \emph{hard} negatives.

\paragraph{In-batch negatives}
In-batch negatives are the other answers from the batch, including them into the set of non-relevant answers is extremely efficient as their representations are already computed.

\paragraph{Hard negatives}
Hard negatives are close but incorrect answers to the questions. Including them improves the performance of retrieval models \cite{karpukhin-etal-2020-dense, DBLP:journals/corr/abs-2007-00808}.
Hard negatives can either come from a standard retrieval system such as BM25, or an earlier iteration of the dense model \cite{DBLP:journals/corr/abs-2007-00808,oguz2021domainmatched}. 
The structure of our dataset, pages of FAQs, facilitates the search for hard negatives. As an example in Table \ref{example-dataset}, three out of four answers share the term \emph{COVID-19}. The model now has to understand the semantic of sentences instead of matching on shared named entities. By including all the pairs of the same page in the same training batch, we ensure that in-batch negatives act as hard negatives.\footnote{To create our batches of training data, we incrementally augment the batch with pairs of a given page. When the batch size reaches the desired size, we start over with the remaining pairs.}

\paragraph{Multilingual}
Although XLM-Roberta is multilingual, we do not expect the model to perform cross-lingual retrieval (i.e. using one language for the query and another for the answer). We make sure that each batch is composed of pairs from the same language. This increases the difficulty of the task. Otherwise, the model could rely on the language of answers as a differentiating factor.

\begin{figure}
    \centering 
    \includegraphics{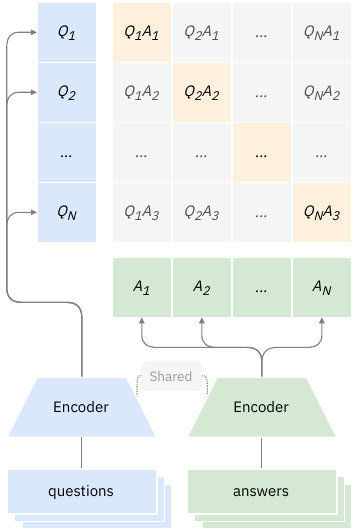} 
        \caption{
        Diagram of our architecture. A shared encoder encodes the questions and the answers independently. Each question's representation (vector) is compared to each answer's representation from the same batch using a dot-product.
    }
    \label{model}
\end{figure}

\section{Experiments}

\begin{table}
\centering
\begin{tabular}{lccc}
\hline
\textbf{Language} & \textbf{TF-IDF} & \textbf{USE} & \textbf{XLM-R} \\ \hline
English            & 63.8            & 64.2         & \textbf{82.5}        \\
German            & 58.0            & 61.8         & \textbf{81.3}        \\
Spanish           & 60.5            & 61.6         & \textbf{81.7}        \\
French            & 60.6            & 62.4         & \textbf{80.7}        \\
Italian           & 58.6            & 55.7         & \textbf{74.7}        \\
Dutch             & 62.9            & 59.6         & \textbf{81.2}        \\
Portuguese        & 55.8            & 56.2         & \textbf{77.4}        \\
Turkish           & 59.2            & 55.7         & \textbf{78.8}        \\
Russian           & 59.2            & 63.1         & \textbf{82.1}        \\
Polish            & 59.2            & 59.9         & \textbf{85.2}        \\
Indonesia         & 71.3            & 62.1         & \textbf{88.5}        \\
Norwegian         & 58.9            & 36.9         & \textbf{83.1}        \\
Swedish           & 59.3            & 36.7         & \textbf{83.3}        \\
Danish            & 64.0            & 42.1         & \textbf{82.7}        \\
Vietnamese        & 73.3            & 43.2         & \textbf{81.2}        \\
Finnish           & 53.5            & 33.2         & \textbf{82.6}        \\
Romanian          & 57.8            & 40.7         & \textbf{83.2}        \\
Czech             & 48.2            & 26.9         & \textbf{69.0}        \\
Hebrew            & 61.5            & 26.5         & \textbf{83.6}        \\
Hungarian         & 38.1            & 28.6         & \textbf{69.7}        \\
Croatian          & 58.1            & 41.4         & \textbf{83.6}        \\ \hline
\end{tabular}
\caption{MRR on MFAQ using various methods. XLM-RoBERTa is a single model trained on all languages at once.}
\label{tab:short-multilingual-results}
\end{table}

In this section, we evaluate the retrieval performance of our model on MFAQ. In all our experiments, we use three metrics to evaluate the performance: precision-at-one (P@1), mean reciprocal rank (MRR), and recall-at-5 (R@5). For space reasons, we only report on MRR in the main text, the full results are available in the annex. We used the same parameters for all experiments unless mentioned otherwise.\footnote{We used a batch size of 800, sequences were limited to 128 tokens (capturing the entirety of ~90\% of the dataset), an Adam optimizer with a learning rate of 0,0001 (warmup of 1000 steps). Dropout of 25\%.} We insert a special token <question> before questions to let the shared encoder know it is encoding a question. Answers are respectively prepended with <answer>. All of our experiments use a subset of the training set: only one page per domain as this technique achieves higher results. Refer to Section \ref{section:subset-training-data} for more information.

We start by studying the performance of multilingual models, then compare it against monolingual models.

\subsection{Multilingual}
We present in Table \ref{tab:short-multilingual-results} a summary of the results of our multilingual training. The model is trained concurrently on the 21 available languages. XLM-RoBERTa achieves a higher MRR on every language compared to the baselines. Low resource languages achieve a relatively high score which could indicate inter-language transfer learning.

\subsection{Monolingual}
Next, we attempt to study if a collection of monolingual models are better suited than a single monolingual model. We use language-specific BERT-like models for each language. The list of BERT models per language is available in the annex. We followed the same procedure as described in Section \ref{xlm-roberta}, except for the encoder which is language-specific.

We limited our study of monolingual models to the ten largest languages of MFAQ. We choose these languages as they have sufficient training examples, and pre-trained BERT-like models are readily available. To study the performance of monolingual models we train models using the same procedure as described in Section \ref{xlm-roberta} except for the encoder.

The results in Table \ref{tab:short-monolingual-results} indicates that a multilingual model outperforms monolingual models in all cases, except for English. These results indicate that leveraging additional languages is beneficial for the task of FAQ retrieval, especially for languages with fewer resources available.
Interestingly, RoBERTa slightly beats DPR in English. This underperformance could be explained by the difference in batch size. Because of the dual encoder nature of DPR, we had to reduce the batch size to 320 compared to 800 for RoBERTa.

\begin{table}
\centering
\resizebox{\columnwidth}{!}
{
\begin{tabular}{lcc}
\hline
\textbf{Language} & \textbf{Monoligual} & \textbf{Multilingual} \\ \hline
En. (DPR)     & 80.5                & 82.5                  \\
En. (RoBERTa) & \textbf{82.9}       & 82.5                  \\
German            & 81.1                & \textbf{81.3}         \\
Spanish           & 78.0                & \textbf{81.7}         \\
French            & 71.0                & \textbf{80.7}         \\
Italian           & 64.1                & \textbf{74.7}         \\
Dutch             & 70.4                & \textbf{81.2}         \\
Portuguese        & 68.4                & \textbf{77.4}         \\
Turkish           & 76.1                & \textbf{78.8}         \\
Russian           & 71.6                & \textbf{82.1}         \\
Polish            & 73.9                & \textbf{85.2}         \\ \hline
\end{tabular}
}
\caption{MRR of monolingual models versus a single multilingual model. The multilingual model outperforms monolingual models in all languages, except for English.}
\label{tab:short-monolingual-results}
\end{table}

\subsection{Cross-lingual}
Our training procedure ensures that the model never has to use language as a cue to select the appropriate answer. Batches of training data all share the same language. We tested the cross-lingual retrieval capabilities of our multilingual model by translating the queries to English while keeping the answers in the original language. The French performance drops from 80.7 to 78.2, which is still better than the unsupervised baselines. The full results are presented in Table \ref{tab:cross-lingual}.

\begin{table}
\centering
\begin{tabular}{lcc}
\hline
\textbf{Language} & \textbf{Cross-lingual} & \textbf{Multilingual} \\ \hline
French            & 78.2                   & 80.7                  \\
Hungarian         & 65.9                   & 69.7                  \\
Croatian          & 71.2                   & 83.6                  \\ \hline
\end{tabular}
\caption{MRR results of our cross-lingual analysis. Questions were translated to English while answers remained in the original language.}
\label{tab:cross-lingual}
\end{table}

A subsection{Subset of training data} \label{section:subset-training-data}
We tested the effect of limiting the number of FAQ pages per domain by limiting the training set to one page per web domain. Using this technique, we achieved an average MRR of 80.8 while using all the training data to reach an average MRR of 76.7. Filtering the training set flattens the topic distribution and better matches the validation set. Another possible approach is to randomly select a given page from a domain at each epoch. This technique would act as a natural regularization. This is left for future work. 

\section{Qualitative analysis}
In this section, we dive into the model's predictions and try to understand why and where it goes wrong. We do so by focusing on a single FAQ page from the admission center of the \emph{Tepper School of Business}.\footnote{It was the first page with less than 25 pairs to end with a \emph{.edu} extension.} The FAQs are displayed in Table \ref{tab:tepper} in the annex. The multilingual model is correct on 74.07\% of the pairs, with an MRR of 85.49. Our qualitative analysis reveals that the model is bad at coreference resolution and depends on keywords for query-answer matching.

\paragraph{Coreference Resolution} The model makes a wrong prediction in question 4 \emph{Can the GMAT or GRE requirement be waived? No, these test scores are required}. The model is unable to guess that \emph{test scores} refer to GMAT or GRE. By changing the answer to \emph{No, the GMAT or GRE scores are required}, the model correctly picks the right answer.

\paragraph{Paraphrase} To study if the model is robust to paraphrasing, we change question 1 from \emph{« Are the hours flexible enough for full-time working adults? »} to \emph{« Is it manageable if I already have a full-time job? »} In this case, the model correctly identifies the right answer. However, if we remove the \emph{full-time} cue, the right answer only arrives in the fourth position. Next, we look at question 15, the model makes a wrong prediction as \emph{opportunities} is not mentioned in the answer. Changing the question to \emph{« It’s a part-time online program, but are there any on-campus [experiences|activities] for students? »} leads to a correct prediction.\footnote{replacing \emph{opportunities} with \emph{events} does not work.}

\paragraph{Keyword search} We replace some questions with a single keyword. We reduced questions 12, 14, 16 and 20 to \emph{cohort}, \emph{payment plan}, \emph{soldier veteran} and \emph{technical requirements}. In all cases, the model guessed correctly, showing the model can do a keyword-based search.

Although it can cope with some synonyms (activities - experiences), this qualitative analysis shows our model is overly reliant on keywords for matching questions and answers. Further research on adversarial training of FAQ retrieval is needed.

\section{Future Work}
Important non-Indo-European languages such as Chinese, Hindi, or Japanese are missing from this dataset. Future work is needed to improve data collection in these languages.
Second, we did not evaluate the model on a real-life FAQ retrieval dataset (with user queries). Future work is needed to see if our model can perform question-to-question retrieval, or if it needs further training to do so.
A linguistic study could analyze the model's strengths and weaknesses by studying the model's performance by type of questions, answers, and entities.

\section{Conclusion}
In this work, we presented the first multilingual dataset of FAQs publicly available. Its size and breadth of languages are significantly larger than other datasets available. While language-specific BERT-like models can be applied to the task of FAQ retrieval, we showed it is beneficial to use a multilingual model and train on all languages at once. This method of training outperforms all monolingual models, except for English. Our qualitative analysis reveals our model is overly reliant on keywords to match questions and answers.

\section{Acknowledgements}
This research received funding from the Flemish Government under the “Onderzoeksprogramma Artificiële Intelligentie (AI) Vlaanderen” programme.

\bibliography{anthology,custom}
\bibliographystyle{acl_natbib}

\clearpage
\appendix

\begin{landscape}
\begin{table}
\begin{tabular}{|l|cll|ccc|ccc|ccc|ccc|ccc|}
\hline
\multirow{2}{*}{\textbf{Language}} &
  \multicolumn{3}{c|}{\textbf{Random}} &
  \multicolumn{3}{c|}{\textbf{TF-IDF}} &
  \multicolumn{3}{c|}{\textbf{USE}} &
  \multicolumn{3}{c|}{\textbf{\begin{tabular}[c]{@{}c@{}}XLM-Roberta\\ (1 page per domain)\end{tabular}}} &
  \multicolumn{3}{c|}{\textbf{\begin{tabular}[c]{@{}c@{}}XLM-RoBERTa\\ (full training set)\end{tabular}}} &
  \multicolumn{3}{c|}{\textbf{Monolingual}} \\ \cline{2-19} 
           & P@1  & MRR  & R@5  & P@1  & MRR  & R@5  & P@1  & MRR  & R@5  & P@1  & MRR           & R@5  & P@1  & MRR  & R@5  & P@1  & MRR           & R@5  \\ \hline
English    & 5.9  & 18.9 & 29.7 & 53.9 & 63.8 & 79.8 & 52.6 & 64.2 & 83.0 & 74.9 & 82.5          & 93.5 & 72.5 & 80.7 & 92.5 & 75.6 & \textbf{82.9} & 93.5 \\
German     & 5.8  & 18.3 & 28.8 & 48.0 & 58.0 & 74.8 & 49.8 & 61.8 & 81.6 & 73.0 & \textbf{81.3} & 93.7 & 69.0 & 78.2 & 92.0 & 72.5 & 81.1          & 93.6 \\
Spanish    & 7.3  & 22.3 & 36.7 & 49.2 & 60.5 & 78.9 & 49.3 & 61.6 & 82.0 & 72.8 & \textbf{81.7} & 94.5 & 68.9 & 78.6 & 93.2 & 68.6 & 78.0          & 92.3 \\
French     & 6.1  & 19.4 & 30.3 & 49.9 & 60.6 & 78.0 & 50.2 & 62.4 & 82.5 & 72.2 & \textbf{80.7} & 93.3 & 68.9 & 78.0 & 91.8 & 60.5 & 71.0          & 87.4 \\
Italian    & 5.2  & 16.8 & 26.2 & 49.0 & 58.6 & 74.4 & 44.1 & 55.7 & 75.2 & 65.9 & \textbf{74.7} & 88.6 & 60.2 & 70.2 & 85.6 & 53.3 & 64.1          & 81.6 \\
Dutch      & 5.3  & 17.3 & 26.5 & 53.2 & 62.9 & 78.6 & 47.7 & 59.6 & 79.2 & 73.0 & \textbf{81.2} & 93.2 & 69.8 & 78.6 & 91.7 & 60.1 & 70.4          & 86.8 \\
Portuguese & 5.3  & 17.0 & 26.6 & 45.5 & 55.8 & 72.7 & 44.1 & 56.2 & 75.8 & 68.5 & \textbf{77.4} & 90.3 & 65.6 & 74.8 & 88.8 & 58.3 & 68.4          & 84.6 \\
Turkish    & 6.2  & 19.0 & 31.0 & 49.3 & 59.2 & 75.2 & 43.5 & 55.7 & 76.4 & 70.2 & \textbf{78.8} & 91.6 & 64.5 & 74.5 & 89.6 & 65.7 & 76.1          & 91.4 \\
Russian    & 7.1  & 21.7 & 35.7 & 48.5 & 59.2 & 76.7 & 49.8 & 63.1 & 83.8 & 73.5 & \textbf{82.1} & 94.4 & 68.9 & 78.7 & 93.1 & 61.0 & 71.6          & 88.3 \\
Polish     & 6.1  & 19.4 & 30.4 & 49.9 & 59.2 & 74.9 & 47.4 & 59.9 & 81.2 & 77.6 & \textbf{85.2} & 96.0 & 73.2 & 81.6 & 94.6 & 64.0 & 73.9          & 89.8 \\
Indonesian & 8.0  & 23.8 & 40.1 & 61.8 & 71.3 & 86.4 & 49.3 & 62.1 & 83.5 & 82.2 & \textbf{88.5} & 97.2 & 76.6 & 84.3 & 95.4 & -    & -             & -    \\
Norwegian  & 5.5  & 17.8 & 27.5 & 48.8 & 58.9 & 75.9 & 25.3 & 36.9 & 57.4 & 76.5 & \textbf{83.1} & 93.7 & 70.6 & 79.4 & 93.3 & -    & -             & -    \\
Swedish    & 5.0  & 16.1 & 25.0 & 49.1 & 59.3 & 76.1 & 25.7 & 36.7 & 56.3 & 75.6 & \textbf{83.3} & 94.6 & 72.0 & 80.5 & 93.1 & -    & -             & -    \\
Danish     & 5.6  & 18.1 & 27.8 & 54.3 & 64.0 & 80.6 & 30.4 & 42.1 & 63.3 & 75.4 & \textbf{82.7} & 92.9 & 71.5 & 79.5 & 91.8 & -    & -             & -    \\
Vietnamese & 11.3 & 30.6 & 56.6 & 62.7 & 73.3 & 90.6 & 28.4 & 43.2 & 70.7 & 73.9 & \textbf{81.2} & 92.5 & 69.8 & 78.3 & 91.6 & -    & -             & -    \\
Finnish    & 5.6  & 18.3 & 28.2 & 43.9 & 53.5 & 70.0 & 21.8 & 33.2 & 53.7 & 74.9 & \textbf{82.6} & 93.9 & 68.5 & 76.8 & 89.7 & -    & -             & -    \\
Romanian   & 6.4  & 20.3 & 32.1 & 48.6 & 57.8 & 73.0 & 29.3 & 40.7 & 62.0 & 76.9 & \textbf{83.2} & 91.5 & 66.6 & 75.4 & 88.2 & -    & -             & -    \\
Czech      & 3.8  & 12.9 & 19.0 & 38.3 & 48.2 & 64.0 & 18.3 & 26.9 & 42.4 & 59.6 & \textbf{69.0} & 83.5 & 50.1 & 60.2 & 77.2 & -    & -             & -    \\
Hebrew     & 8.6  & 25.1 & 42.8 & 49.3 & 61.5 & 81.3 & 14.5 & 26.5 & 50.7 & 75.3 & \textbf{83.6} & 95.5 & 68.8 & 78.7 & 93.7 & -    & -             & -    \\
Hungarian  & 4.1  & 13.3 & 20.4 & 30.3 & 38.1 & 52.1 & 21.0 & 28.6 & 41.4 & 60.6 & \textbf{69.7} & 83.7 & 54.1 & 64.1 & 80.5 & -    & -             & -    \\
Croatian   & 4.9  & 15.9 & 24.5 & 49.4 & 58.1 & 73.0 & 32.8 & 41.4 & 56.7 & 78.2 & \textbf{83.6} & 92.6 & 71.8 & 79.4 & 91.4 & -    & -             & -    \\ \hline
\end{tabular}
\caption{Results of our experiments on MFAQ. XLM-RoBERTa (1 page per domain) is consistently better than the rest, except for English where a RoBERTa model achieves a higher MRR.
P@1 = Precision-at-1 (accuracy), MRR = Mean Reciprocal Rank, R@5 = Recall-at-5, One page per domain = subset of the training set.}
\label{tab:all-results-single-domain}
\end{table}
\end{landscape}

% \section{Special tokens}
% The shared encoder uses a special token prepended at the start of sequences to represent questions and answers (<question>, <answer>). According to our analysis, the model make little use of these tokens. We tried removing both <questions> and <answer> or only one of each. In each scenario the change in performance is low. See Table \ref{table-sepcial-tokens} for the details results. Further research is necessary to asses if training without these tokens can boost the performance.

% \begin{table}[]
% \begin{tabular*}{\columnwidth}{@{\extracolsep{\fill}}lccc}
% \hline
% \textbf{Special tokens}                & \textbf{P@1} & \textbf{MRR} & \textbf{R@5} \\ \hline
% None                                   & 74.6         & 82.3         & 93.3         \\
% Only \textless{}answer\textgreater{}   & 74.7         & 82.4         & 93.4         \\
% Only \textless{}question\textgreater{} & 74.8         & 82.4         & 93.4         \\
% \textless{}question\textgreater \&\textless{}answer\textgreater{} & \textbf{74.9} & \textbf{82.5} & \textbf{93.5} \\ \hline
% \end{tabular*}
% \caption{
% Special tokens to represent questions and answers bring little benefits to the overal performance.
% }
% \label{table-sepcial-tokens}
% \end{table}

% \usepackage{graphicx}

\begin{landscape}
\begin{table}
\centering
\tiny
\begin{tabular}{p{0.25cm}p{5.5cm}p{15cm}}
\textbf{ID} &
  \textbf{Question} &
  \textbf{Answer} \\ \hline
0 &
  Are international students eligible for the MSBA program? &
  Yes, international students are eligible   for the MSBA program. Please review the International Applicants page for   specific requirements. \\ \hline
1 &
  Are the hours flexible enough for full-time working adults? &
  Yes, the MSBA program accommodates students working full-time. Required   weekly live sessions, lasting 75 minutes, are held in the evening, and the   three residential components, two strongly recommended and one optional, take   place over weekends. Students complete all other coursework on their own   schedule, but must adhere to deadlines and be prepared to participate in   weekly live sessions. \\ \hline
2 &
  Can I take a course from a third-party provider, like Lynda or Coursera, to prepare for the programming requirements   of this program? &
  Our goal is to make sure that everyone entering the program has the   necessary background to be successful. We strongly recommend that applicants   who feel they need additional preparation in programming languages take a   for-credit course from an accredited two- or four-year institution. \\ \hline
3 &
  Can I transfer credits into the program? &
  No, the Tepper School does not accept transfer credits. \\ \hline
4 &
  Can the GMAT or GRE requirement be waived? &
  No, these test scores are required. \\ \hline
5 &
  Do I have to maintain a certain GPA in the program to graduate? &
  Yes, MSBA degree candidates must maintain a minimum cumulative GPA of 3.0   to graduate. \\ \hline
6 &
  Do you offer the opportunity to preview courses in your program to get a feel for what they are like? &
  Yes we do. To preview one of our courses, please visit our Virtual Class   Visit page. You’ll be able to register to virtually participate in a course   of your choosing. \\ \hline
7 &
  How do I learn more about the online   learning environment? &
  To preview one of our courses, please visit our Virtual Class Visit page.   You’ll be able to view upcoming courses and register to virtually attend a   course of your choosing. \\ \hline
8 &
  How many hours per week should be   dedicated to coursework? &
  Students take two classes at a time and should expect to spend at least   10 hours on each course, or 20 hours total for the week. Coursework includes   live synchronous meetings, assignments, projects, readings, and quizzes. \\ \hline
9 &
  If I need to withdraw from the program,   will I get a refund? &
  If I need to withdraw from the program, will I get a refund? \\ \hline
10 &
  If I’m already proficient in basic   programming and probability/statistics, do I have to take these courses? &
  Yes, the 46-880 Introduction to Probability and Statistics and 46-881   Programming in R and Python courses are required for all MSBA students. These   courses ensure that all students have the necessary skills and knowledge to   succeed in courses that follow. For more information, visit the Curriculum page on our website. \\ \hline
11 &
  Is the MSBA offered exclusively on   campus? &
  No, the MSBA degree is offered only online, with three optional on-campus   experiences. Though they all are optional, we strongly recommend that   students attend the BaseCamp and Capstone Project experiences, which occur at   the beginning and end of the degree program. \\ \hline
12 &
  Is the MSBA program structured in   cohorts? &
  Yes, the part-time, online MSBA is structured in cohorts to optimize   student interaction and success in the program. \\ \hline
13 &
  Is the Tepper School participating in the   Yellow Ribbon Program? &
  Yes, the Tepper School is participating in the Yellow Ribbon   Program. For more information, please visit the Tuition page or contact Mike Danko   at uro-vaedbenefits@andrew.cmu.edu. \\ \hline
14 &
  Is there a Tuition Payment Plan   available? &
  Yes, for more information about a monthly payment plan and debt   minimization services, please review our payment options. \\ \hline
15 &
  It’s a part-time online program, but are   there any on-campus opportunities for students? &
  We have three on-campus experiences. The first is an orientation   basecamp, where the students are introduced to the program, interact with   faculty, and learn about their cohort. The second, an immersive analytics   experience led by top CMU faculty, takes place mid-program. [...] \\ \hline
16 &
  I’m an active duty soldier/veteran. Am I   eligible for an application fee waiver? &
  Yes, as a GMAC military-friendly business school, we waive the \$125   application fee for active duty U.S. military personnel, veterans and   retirees. Please contact Mike Danko at uro-vaedbenefits@andrew.cmu.edu to   discuss the fee waiver. \\ \hline
17 &
  Must international students come to   campus? &
  We recommend attendance at the on-campus experiences, but students who   are unable to attend may participate remotely and still meet the requirements   of the program. Please note that because the program is delivered online,   enrollment in the MSBA will not qualify students for a student visa to enter   the United States. \\ \hline
18 &
  What are some examples of roles a   graduate could pursue after the program? &
  Business analytics professionals hold a range of positions across sectors   and industries. They have titles such as business intelligence analyst,   operations research analyst, market research analyst and statistician. Other   job titles for these professionals are available here. \\ \hline
19 &
  What are the programming languages that I   should have experience in before applying to the program? &
  Basic programming knowledge in a modern language is required for   admission. You do not need to be familiar with any specific language or build   advanced programming skills before applying to the MSBA program. Your courses   in the program will introduce you to relevant languages and provide hands-on   experience. \\ \hline
20 &
  What are the technical requirements for the MSBA program? &
  All students must have access to the following technologies in order to   participate in the program:      Laptop with the following requirements: - Windows – Intel Core i5 processor or higher; 8GB RAM, 256+ hard drive   capacity - Macintosh [...] \\ \hline
21 &
  What career resources are available for   MSBA students and alumni? &
  The Master’s Career Center helps students develop strategies focused on   their career needs through a variety of services. For example, the career   center hosts workshops and webinars in job search fundamentals, such as   resume writing, interviewing, and networking. [...] \\ \hline
22 &
  What happens if I need to defer starting   or withdraw from the program? &
  Deferrals are granted only if an applicant must complete military service   or has an extreme emergency. Deposits are refunded in these instances.   Students are re-admitted the following year and must submit their deposit   before the deadline for their start date. [...] \\ \hline
23 &
  What is the average Quant and Verbal   scores for the GRE and GMAT? &
  There is no average score expectation. The test scores are simply one   component of the multifaceted admissions process that we consider when making   an admissions decision. \\ \hline
24 &
  What separates the Tepper School of   Business’ online MSBA program from other MSBA programs, either online or   on-campus? &
  The Tepper School of Business is globally renowned for its analytical   approach to business problem solving. It is an integral part of Carnegie   Mellon University, a top-tier research university that has become the center   for disciplines including data science, robotics, business intelligence and   additive manufacturing. [...] \\ \hline
25 &
  What time(s) do the synchronous sessions   take place? &
  The weekly live sessions are in the evening (U.S. Eastern Time) and   typically last 75 minutes. \\ \hline
26 &
  What types of financial aid or   scholarships are available to online students? &
  Students may be eligible to take out federal and/or private education   loans to cover tuition and other education-related costs. Please view our   Tuition page for details. At this time, the Tepper School does not provide scholarships for the MSBA   program. \\ \hline
\end{tabular}
\caption{FAQ pairs from the Tepper School of Business}
\label{tab:tepper}
\end{table}
\end{landscape}

\end{document}